\documentclass[journal]{IEEEtran}
\ifCLASSINFOpdf
\else
\fi
\usepackage{amsfonts}
\usepackage{amsmath}
\usepackage{amssymb}
\usepackage{graphicx}
\usepackage{algorithm}
\usepackage{algpseudocode}
\usepackage{array}
\usepackage{booktabs}
\usepackage{makecell}
\usepackage{cite}
\usepackage{color}
\usepackage{soul}
\usepackage{bm}
\usepackage{eufrak}
\usepackage{multirow}
\soulregister\ref7
\soulregister\cite7

\newtheorem{definition}{Definition}
\newcommand{\bea}{\begin{eqnarray}}
\newcommand{\eea}{\end{eqnarray}}

\renewcommand{\baselinestretch}{0.963}

\ifCLASSOPTIONcompsoc
\usepackage[caption=false,font=normalsize,labelfon
t=sf,textfont=sf]{subfig}
\else
\usepackage[caption=false,font=footnotesize]{subfi
g}
\fi

\begin{document}
%
\title{Optimization of Directional Landmark Deployment for Visual Observer on \emph{SE}(3)}
%
%
%

\author{Zike~Lei,
        Xi~Chen,~\IEEEmembership{Member,~IEEE,}
        Ying~Tan,~\IEEEmembership{Senior~Member,~IEEE,}
        Xiang~Chen,~\IEEEmembership{Member,~IEEE,}
        Li~Chai,~\IEEEmembership{Member,~IEEE}
\thanks{*This work was supported under the National Natural Science Foundation
of China under Grant 61703315, 61625305.}%
\thanks{Zike Lei is with the School of Information Science and Engineering, Wuhan University of Science and Technology, Hubei, 430081 China (e-mail: leizike@wust.edu.cn).}
\thanks{Xi Chen is with the Engineering Research Center of Metallurgical Automation and Measurement Technology, Wuhan University of Science and Technology, Hubei 430081, China (e-mail: chenxi\_99@wust.edu.cn) (corresponding author).}
\thanks{Ying Tan is with the School of Electrical, Mechanical and Infrastructure Engineering, The University of Melbourne, VIC 3010 Australia (e-mail: yingt@unimelb.edu.au).}
\thanks{Xiang Chen is with the Department of Electrical and Computer Engineering, University of Windsor, Ontario, N9B 3P4 Canada (e-mail: xchen@uwindsor.ca).}
\thanks{Li Chai is with the College of Control Science and Engineering, Zhejiang University, Hangzhou, 310027, China (e-mail: chaili@zju.edu.cn).}
}
\maketitle

\begin{abstract}

An optimization method is proposed in this paper for novel deployment of given number of directional landmarks (location and pose) within a given region in the 3-D task space. This new deployment technique is built on the geometric models of both landmarks and the monocular camera. In particular, a new concept of Multiple Coverage Probability (MCP) is defined to characterize the probability of at least $\bm n$ landmarks being covered simultaneously by a camera at a fixed position. The optimization is conducted with respect to the position and pose of the given number of landmarks to maximize MCP through globally exploration of the given 3-D space. By adopting the elimination genetic algorithm, the global optimal solutions can be obtained, which are then applied to improve the convergent performance of the visual observer on \emph{SE}(3) as a demonstration example. Both simulation and experimental results are presented to validate the effectiveness of the proposed landmark deployment optimization method.

\end{abstract}

\begin{IEEEkeywords}
Landmark deployment, multiple coverage probability, visual observer on \emph{SE}(3), robotic vision.
\end{IEEEkeywords}

%
\IEEEpeerreviewmaketitle

\section{Introduction}
\label{section1}
%
%
%
%

\IEEEPARstart{A}{s} a kind of field sensor, camera sensors are selected in the observer as they have 
the advantages of rich information, moderate cost, low power consumption, light weight, which have been widely used in various visual tasks \cite{mavrinac2013modeling,lei2020modeling,lei2021radial} and localization-related tasks \cite{hua2015gradient,wang2014simple,mezouar2002path}. In these tasks. various landmarks have been used in indoor localization tasks in 2-D space or 3-D space \cite{zhang2015localization,nazemzadeh2017indoor,sani2017automatic}, along with camera to localize the sensor itself. Visual observers are needed to estimate the position and orientation of the robotic system on Special Euclidean group SE(3) from the measurements from camera \cite{mahony2008nonlinear,cattaneo2020global}.

Many visual observers on \emph{SE}(3) have been proposed. Among them, a landmark-based observer on \emph{SO}(3) are original proposed by Mahony in \cite{mahony2008nonlinear} to estimate the orientation of moving robots, and then extended to the observer on \emph{SE}(3) in \cite{hua2015gradient, zhang2021damping}. Although existing visual observers work well, they all assume that all landmarks can be captured by sensors moving with the robot. 
This assumption does not always hold in engineering applications. For examples, when cameras are used to detect landmarks, due to the geometric limitation of the camera frustum, some landmarks might not be identified when cameras are moving along with the robot. This will lead to a poor performance of the visual observer in terms of localization as shown in the illustrative example. 

In order to improve the performance of the existing visual observers on \emph{SE}(3), this work exploits how to optimally deploy the given number of landmarks in a given 3-D space. In particular, to exploit the flexibility coming from orientation of landmarks, directional landmarks are considered in this work. Furthermore, many landmarks deployment techniques have been proposed to improve in-door localization accuracy. Some techniques are quite general as they can be applied to a large class of sensors and landmarks as \cite{chen2006practical,benbadis2007exploring,perez2013optimal,wang2019landmark}.
However, these general solution cannot always produce the optimal solutions for the special cases without taking models of sensors and landmark into consideration. 
2-D camera models have been considered in \cite{nazemzadeh2016optimal,magnago2017nearly,magnago2018optimal} without dealing with the orientation of landmarks, while the various landmark models are used in \cite{nguyen2006simple, ji2018three} without specifying the types of sensors used in deployment. To the best of authors' knowledge, there is no landmark deployment technique, which considers 3-D camera models and directional landmarks models simultaneously in literature. The challenge of this problem is how to deal with
the relationship between the geometry model of the camera
and landmarks, and how to evaluate the performance of the
landmark deployment when constraints coming from these geometry models exist.

This paper reformulates the landmark deployment as an optimization problem, which maximizes the chances of at least $n$ landmarks being covered simultaneously by a camera at a fixed position. Consequently, the concept of multiple coverage probability (MCP) is proposed. Exploring MCP at each position over a given 3-D space results in a global optimization problem with respect to landmark position and orientation. The elimination genetic algorithm (EGA) is modified from the standard genetic algorithm (SGA) used in \cite{lei2021radial} to solve this global optimization problem. In order to show the effectiveness of the proposed landmark deployment technique, the obtained optimal landmark deployment is used in the existing visual observers on \emph{SE}(3) in \cite{zhang2021damping} as an example due to it is a relatively new research on visual observers. Both simulation results and experiment results using an unmanned aerial vehicle (UAV) equipped with a camera show the significant performance improvement in terms of the observer accuracy, compared with uniform landmark distribution and random landmark distribution, which are widely used in landmark setting. 

The organization of the remaining parts of this paper is as follows: the motivation and some preliminary knowledge are presented in Section \ref{section2}; the multiple coverage probability is proposed in detail in Section \ref{section3}; in Section \ref{section4}, the landmark deployment is formulated, followed by the introduction of  elimination genetic algorithm to find its optimal solution. Section \ref{section5} shows the simulation and experiment results to validate the performance improvement of the visual observer; Section \ref{section6} concludes the paper.

\emph{Notations:} $\mathbb{R}^n$ denotes a $n$-dimensional Euclidean space, $\mathbb{R}^+$ denotes positive real number, and $\mathbb{N}^+$ represents a positive integer. \emph{SO}(3) denotes the Special Orthogonal Group and \emph{SE}(3) denotes the Special Euclidean Group in the 3-D space. Let $\mathfrak{se}$(3) be the Lie algebra of \emph{SE}(3). $\|\cdot\|$ stands for the Euclidean norm and $\|\cdot\|_\mathrm{F}$ denotes the Frobenius norm of a vector.

\section{Motivation and Preliminaries}
\label{section2}
This section introduces the need of landmark deployment in visual observer design on \emph{SE}(3) by a motivating example.
The preliminaries of camera models and landmark models needed are also provided.

\subsection{A Motivating Example}
Zhang proposed a robust observer on \emph{SE}(3) in \cite{zhang2021damping} to estimate the position and orientation of the onboard sensor (such as camera) from the captured landmarks information. This subsection will briefly give the formula of the visual observer, and the details can be found in references.

The robotic motion of a moving camera can be represented mathematically with Lie group/algebra of the following form:
\begin{equation}
\label{system state}
        \dot{X} = XU,
\end{equation}
where $X \in \emph{SE}(3)$ is the state and $U \in \mathfrak{se}(3)$ is the input. The visual observer in \cite{zhang2021damping} is formulated as
\begin{equation}
\label{X_hat_dot}
\dot{\hat{X}} = \hat{X}(U-\varepsilon-\mathbb{P}(k_0I_4Q(\hat{Y}-Y))), 
\end{equation}
where $\varepsilon$ has the following form:
\begin{equation}
      \varepsilon = \hat{X}^{-1}grad_{\hat{X}}f(\hat{X},X)
\end{equation}
Let $C_i$ denotes the position vector of the $i^{th}$ landmark in the initial frame, the output of the system is written as
\begin{equation}
\label{multiplelandmark}
Y =  \sum_{i=1}^{K} X^{-1}C_i,
\end{equation}
\begin{equation}
\label{observeroutput}
\hat{Y} = \sum_{i=1}^{K} \hat{X}^{-1}C_i,
\end{equation}
and the cost function of the visual observer is
\begin{equation}
\label{f_without}
f(\hat{X},X) = \frac{1}{2}k_i\sum_{i=1}^K\|(\hat{X}^{-1} - X^{-1})C_i\|_\mathrm{F}^2.
\end{equation}

More details can be found in \cite{zhang2021damping}. It is worth noting that this visual observer assumes that all landmarks can be captured by the camera, without considering the geometric limitations of the camera. 
In engineering applications, the camera cannot always capture all landmarks. As the camera pose changes over time, the captured number of landmark will also change. 

\begin{figure}[!t]
\centering
\includegraphics[scale=1]{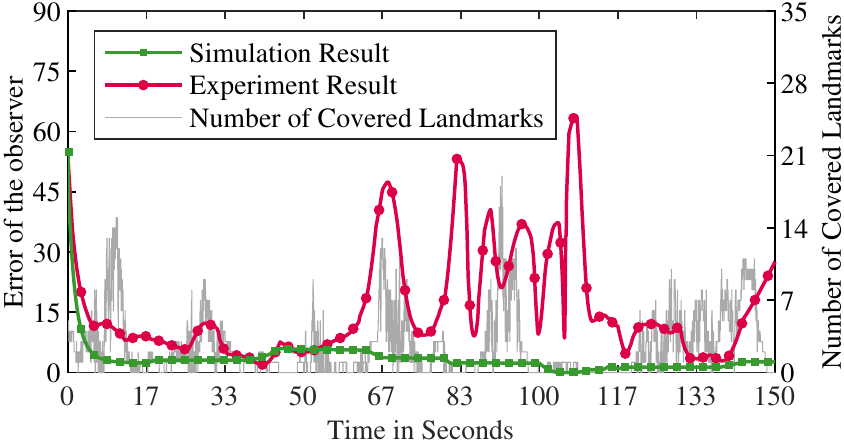}
\caption{Comparison of observer performance with and without camera model.}
\label{observer_compare}
\end{figure}

The visual observer (\ref{X_hat_dot}) is implemented by using a camera on a UAV system with an indoor space $320\times 320\times170$ cm. 
The camera on UAV is used to perform visual observer task. 80 landmarks are randomly distributed on four walls (see more details in Section \ref{section5d}).  The comparison between the simulation result (assuming that the camera can cover all landmarks) and the experiment result in term of estimation error $\|X-\hat{X}\|_F^2$ is shown in Fig. \ref{observer_compare}.  The number of landmarks that can be observed by the camera during the experiment is also presented in Fig. \ref{observer_compare}. For example, at some time instants, the camera can cover less than $8(10\%)$ landmarks, which greatly impacts the performance of the visual observer. This indicates that the position and orientation of landmarks play an important role in the visual observer on $\emph{SE}(3)$ in (\ref{X_hat_dot}), which motivates the work of landmark deployment.

\subsection{Frame Transformation}
\begin{figure}[!t]
\centering
\includegraphics[scale=1]{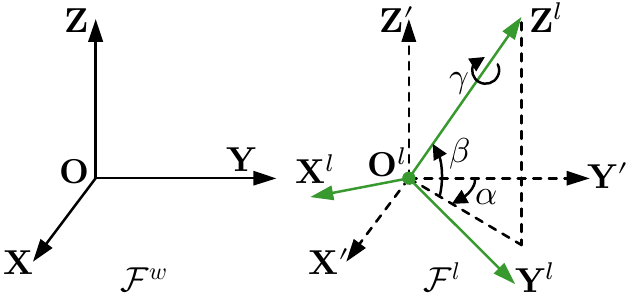}
\caption{World frame and local frame.}
\label{world_local_frame}
\end{figure}

In this paper, two types of frames are defined, including the world frame and the local frame. As shown in Fig. \ref{world_local_frame}, the world frame $\mathcal{F}^w$ is represented by axis $\mathbf{XYZ}$ with $\mathbf{O}$ being its origin, the local frame $\mathcal{F}^l$ is represented by axis $\mathbf{X}^l\mathbf{Y}^l\mathbf{Z}^l$ with $\mathbf{O}^l$ being its origin. The axis $\mathbf{X}'\mathbf{Y}'\mathbf{Z}'$ indicated by the dotted line is the reference axis parallel to $\mathbf{XYZ}$.

Let $\mathbf{c}_l\in\mathbb{R}^6$ denote the coordinates of $\mathcal{F}^l$:
\begin{equation*}
\label{c_l}
\mathbf{c}_l = [\begin{IEEEeqnarraybox*}[][c]{,c/c,}
\boldsymbol{\varsigma}_l^\mathrm{T} & \boldsymbol{\vartheta}_l^\mathrm{T}
\end{IEEEeqnarraybox*}]^\mathrm{T}
= [\begin{IEEEeqnarraybox*}[][c]{,c/c/c/c/c/c,}
x_l & y_l & z_l & \alpha & \beta & \gamma
\end{IEEEeqnarraybox*}]^\mathrm{T},
\end{equation*}
where the position component
$
\boldsymbol{\varsigma}_l = [\begin{IEEEeqnarraybox*}[][c]{,c/c/c,}
x_l & y_l & z_l
\end{IEEEeqnarraybox*}]^\mathrm{T}\in\mathbb{R}^3
$
denotes the coordinates of $\mathbf{O}^l$ in $\mathcal{F}^w$, and the orientation component
$
\boldsymbol{\vartheta}_l = [\begin{IEEEeqnarraybox*}[][c]{,c/c/c,}
\alpha & \beta & \gamma
\end{IEEEeqnarraybox*}]^\mathrm{T}\in\mathbb{R}^3
$
denotes the orientation of $\mathcal{F}^l$, where $\alpha\in[-\pi,\pi)$ is the yaw angle, $\beta\in[-\pi/2,\pi/2]$ is the pitch angle, and $\gamma\in[-\pi,\pi)$ is the roll angle of $\mathcal{F}^l$ measured in $\mathcal{F}^w$. For a point
$
\mathbf{s} = [\begin{IEEEeqnarraybox*}[][c]{,c/c/c,}
x & y & z
\end{IEEEeqnarraybox*}]^\mathrm{T}\in\mathbb{R}^3
$ in $\mathcal{F}^w$, let its coordinate in $\mathcal{F}^l$ be
$
\mathbf{s}^l = [\begin{IEEEeqnarraybox*}[][c]{,c/c/c,}
x^l & y^l & z^l
\end{IEEEeqnarraybox*}]^\mathrm{T}\in\mathbb{R}^3
$
, then the following relationship holds \cite{zhang2000flexible}:
\begin{equation}
\label{s^l}
\mathbf{s}^l = \mathbf{R}(\mathbf{s}-\boldsymbol{\varsigma}_l),~~~~~~
\mathbf{R} = \mathbf{R}^{\gamma}\mathbf{R}^{\beta}\mathbf{R}^{\alpha}\left[\begin{IEEEeqnarraybox*}[][c]{,c/c/c,}
1 & \hspace{2pt}0\hspace{2pt} & 0 \\
\hspace{2pt}0\hspace{2pt} & 0 & -1 \\
0 & 1 & 0
\end{IEEEeqnarraybox*}\right]\!,
\end{equation}
where $\mathbf{R}\in \text{\emph{SO}(3)}$ is the rotation transformation matrix from $\mathcal{F}^w$ to $\mathcal{F}^l$, and
\begin{align*}
\label{R^abg}
\mathbf{R}^{\alpha} &= \left[\begin{IEEEeqnarraybox*}[][c]{,c/c/c,}
\cos{\alpha} & \hspace{3pt}0\hspace{3pt} & -\sin{\alpha} \\
0 & 1 & 0 \\
\sin{\alpha} & 0 & \cos{\alpha}
\end{IEEEeqnarraybox*}\right]\!,~
\mathbf{R}^{\beta} = \left[\begin{IEEEeqnarraybox*}[][c]{,c/c/c,}
1 & 0 & 0 \\
\hspace{3.25pt}0\hspace{3.25pt} & \cos{\beta} & \sin{\beta} \\
0 & -\sin{\beta} & \cos{\beta}
\end{IEEEeqnarraybox*}\right]\!, \\
\mathbf{R}^{\gamma} &= \left[\begin{IEEEeqnarraybox*}[][c]{,c/c/c,}
\cos{\gamma} & \hspace{0.5pt}-\sin{\gamma}\hspace{0.5pt} & \hspace{3.2pt}0\hspace{3.2pt} \\
\sin{\gamma} & \cos{\gamma} & 0 \\
0 & 0 & 1
\end{IEEEeqnarraybox*}\right]\!
\end{align*}
are the rotation transformation matrices from $\mathcal{F}^w$ to $\mathcal{F}^l$ for yaw, pitch and roll, respectively.

\subsection{Landmark Model}

A detailed model of the plate landmark is described for landmark deployment. Let $l_k$ be the $k^{th}$ plate landmark defined in $\mathcal{F}^w$ ($k=1,2,\cdots,K$), with $K\in\mathbb{N}^+$ being the total number of landmarks. For each landmark, a local frame is defined on it. The origin of the frame is defined as the geometric center of the landmark, and $\mathbf{Z}^l$-axis is along the front face normal direction of the landmark. Let the directional point $\mathbf{L}_k\in\mathbb{R}^6$ denotes the coordinate of the unit normal vector of $l_k$:
\begin{equation*}
\label{L_k}
\mathbf{L}_k = [\begin{IEEEeqnarraybox*}[][c]{,c/c,}
\mathbf{s}^\mathrm{T}_k & \boldsymbol{\theta}^\mathrm{T}_k
\end{IEEEeqnarraybox*}]^\mathrm{T}
= [\begin{IEEEeqnarraybox*}[][c]{,c/c/c/c/c/c,}
x_k & y_k & z_k & \rho_k & \eta_k & \mu_k
\end{IEEEeqnarraybox*}]^\mathrm{T},
\end{equation*}
where
$
\mathbf{s}_k = [\begin{IEEEeqnarraybox*}[][c]{,c/c/c,}
x_k & y_k & z_k
\end{IEEEeqnarraybox*}]^\mathrm{T}\in\mathbb{R}^3
$
is the position component and
$
\boldsymbol{\theta}_k = [\begin{IEEEeqnarraybox*}[][c]{,c/c/c,}
\rho_k & \eta_k & \mu_k
\end{IEEEeqnarraybox*}]^\mathrm{T}\in\mathbb{R}^3
$
is the orientation component. Then the front face of normal direction $\mathbf{n}_k$ of $l_k$ can be characterized by
\begin{equation}
\label{n_k}
\mathbf{n}_k = [\begin{IEEEeqnarraybox*}[][c]{,c/c/c,}
-\sin{\rho_k}\cos{\eta_k} & \;\cos{\rho_k}\cos{\eta_k} & \;-\sin{\eta_k}
\end{IEEEeqnarraybox*}]^\mathrm{T}
\end{equation}
in $\mathcal{F}^w$.

\subsection{Camera Model}
A pinhole camera is used to obtain visual information. The frustum of the camera is defined as a truncated cone tangent to the square frustum in \cite{lei2021radial}. The geometric model of the camera is represented by intrinsic parameters and extrinsic parameters \cite{zhang2000flexible}. Intrinsic parameters are the inherent properties of the camera and cannot be adjusted during the execution of the visual task. The intrinsic parameters are shown in Table \ref{intrinsic_parameter}.

\begin{figure}[!t]
\centering
\includegraphics[scale=1]{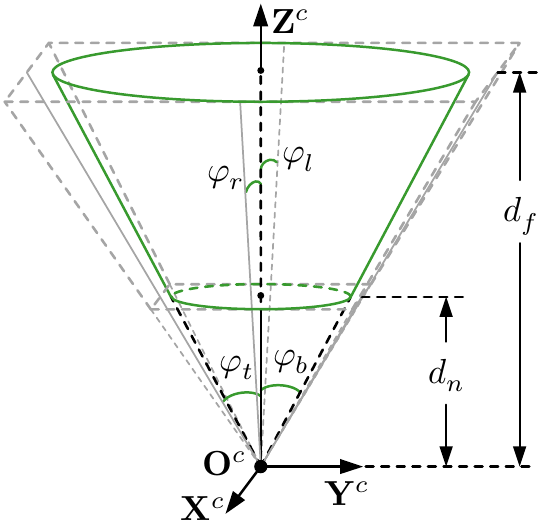}
\caption{The simplified camera model.}
\label{simplified_camera_model}
\end{figure}

A local frame is defined on the camera, called the camera frame $\mathcal{F}^c$. The origin of the frame is defined at the optical center of the camera and $\mathbf{Z}^c$-axis is along the optical axis of the camera. The $\mathbf{X}^c$-axis and $\mathbf{Y}^c$-axis are along the opposite directions of $\mathbf{U}^c$-axis and $\mathbf{V}^c$-axis axes of the photosensitive device, such as the Charge Coupled Device (CCD) or the Complementary Metal Oxide Semiconductor (CMOS), respectively, which form a right-handed frame together with $\mathbf{Z}^c$-axis. The extrinsic parameters of the camera are defined in $\mathcal{F}^w$. Let $\mathbf{c}\in\mathbb{R}^6$ denote the coordinates of the camera:
\begin{equation*}
\label{c}
\mathbf{c} = [\begin{IEEEeqnarraybox*}[][c]{,c/c,}
\boldsymbol{\varsigma}^\mathrm{T} & \boldsymbol{\vartheta}^\mathrm{T}
\end{IEEEeqnarraybox*}]^\mathrm{T}
= [\begin{IEEEeqnarraybox*}[][c]{,c/c/c/c/c/c,}
x_c & y_c & z_c & \alpha_c & \beta_c & \gamma_c
\end{IEEEeqnarraybox*}]^\mathrm{T},
\end{equation*}
where
$
\boldsymbol{\varsigma} = [\begin{IEEEeqnarraybox*}[][c]{,c/c/c,}
x_c & y_c & z_c
\end{IEEEeqnarraybox*}]^\mathrm{T}\in\mathbb{R}^3
$
is the position component and
$
\boldsymbol{\vartheta} = [\begin{IEEEeqnarraybox*}[][c]{,c/c/c,}
\alpha_c & \beta_c & \gamma_c
\end{IEEEeqnarraybox*}]^\mathrm{T}\in\mathbb{R}^3
$
is the orientation component. The camera model is shown in Fig. \ref{simplified_camera_model}. From Equation (\ref{s^l}), the 3-D coordinate $\mathbf{s}_k$ in $\mathcal{F}^w$ can be transformed into the coordinate
$
\mathbf{s}_k^c = [\begin{IEEEeqnarraybox*}[][c]{,c/c/c,}
x_k^c & y_k^c & z_k^c
\end{IEEEeqnarraybox*}]^\mathrm{T}\in\mathbb{R}^3
$ in $\mathcal{F}^c$ by
$
\mathbf{s}_k^c = \mathbf{R}_c(\mathbf{s}_k-\boldsymbol{\varsigma})
$,
where $\mathbf{R}_c\in \text{\emph{SO}(3)}$ is the rotation transformation matrix from $\mathcal{F}^w$ to $\mathcal{F}^c$.

\begin{table}[!t]
\renewcommand{\arraystretch}{1.3}
\caption{Intrinsic Parameters of Camera Model}
\label{intrinsic_parameter}
\centering
\begin{tabular}{p{62pt}l}
\hline\hline
\specialrule{0em}{0pt}{2pt}
Parameter & Description \\
\specialrule{0em}{2pt}{0pt}
\hline
\specialrule{0em}{2pt}{0pt}
$f\in\mathbb{R}^+$ & Lens focal length (mm) \\
$s_u\in\mathbb{R}^+$ & Horizontal pixel dimensions (mm/pixel) \\
$s_v\in\mathbb{R}^+$ & Vertical pixel dimensions (mm/pixel) \\
$
\mathbf{o} \!=\! [\begin{IEEEeqnarraybox*}[][c]{,c/c,}
\!o_u\! & o_v\!
\end{IEEEeqnarraybox*}]^\mathrm{T}\!\!\in\!\mathbb{R}^2
$
& Principle point (pixel) \\
$w\in\mathbb{R}^+$ & Image width (pixel) \\
$h\in\mathbb{R}^+$ & Image height (pixel) \\
$d_a\in\mathbb{R}^+$ & Effective aperture diameter of optical lens (mm) \\
$d_s\in\mathbb{R}^+$ & Focusing distance (mm) \\
$\varphi_t,\varphi_b\in[0,\pi/2)$ & FOV angles to top / bottom image boundary (rad) \\
$\varphi_l,\varphi_r\in[0,\pi/2)$ & FOV angles to left / right image boundary (rad) \\
\specialrule{0em}{2pt}{0pt}
\hline\hline
\end{tabular}
\end{table}

\section{Coverage Probability}
\label{section3}
As shown in the motivating example (see Fig. \ref{observer_compare}), the position and orientation of $K$ landmarks determine the performance of the visual observers on \emph{SE}(3). 
The problem of interests is: for a given set of $K$ landmarks in a given 3-D space, for any location and pose of a camera, we want to find the optimal position and orientation of these landmarks such that the camera can cover the largest possible number of landmarks at any point in this 3-D space. 

For any fixed position of the camera, it has a reachable region, which contains its all possible
orientations. For each landmark, the concept of the coverage spherical cap, which is related to the geometry property of the landmark and the geometry properties of the camera such as resolution, FOV, focus, and occlusion, is introduced.
When the orientation of the camera is within the coverage spherical cap of one landmark, it means that the camera can cover this landmark. 
Consequently, the concept of multiple coverage probability (MCP) is introduced to represent the probability that the camera can cover no less then $n$ landmarks simultaneously. Such a concept plays a key role in constructing the cost function of landmark deployment that will be presented in Section \ref{section4}. 



Let
$
\mathbf{p} = [\begin{IEEEeqnarraybox*}[][c]{,c/c/c,}
x_\mathbf{p} & y_\mathbf{p} & z_\mathbf{p}
\end{IEEEeqnarraybox*}]^\mathrm{T}\in\mathbb{R}^3
$
be any coordinate in the 3-D space. Let $\Omega$ be the set containing all 3-D points in the space of interest. For a given position in $\Omega$, the reachable area refers to all positions that the camera may pass through during the visual task. Suppose the yaw angle and the pitch angle of the camera placed at $\mathbf{p}$ are expressed as $\alpha_{\mathbf{p}}\in[-\pi,\pi), \beta_{\mathbf{p}}\in[-\pi/2,\pi/2]$, respectively. Then the extrinsic parameters of the camera can be written as
\begin{equation*}
\label{c_p}
\mathbf{c}_\mathbf{p} = [\begin{IEEEeqnarraybox*}[][c]{,c/c/c/c/c/c,}
x_\mathbf{p} & y_\mathbf{p} & z_\mathbf{p} & \alpha_\mathbf{p} & \beta_\mathbf{p} & \gamma_\mathbf{p}
\end{IEEEeqnarraybox*}]^\mathrm{T},
\end{equation*}
and the position of $\mathbf{p}$ in $\mathcal{F}^c$ is then expressed as $\mathbf{s}_k^c$. It should be emphasized that since the camera model used in this paper is a truncated cone, the roll angle of the camera will not affect the imaging performance, then $\gamma_\mathbf{p}$ can take any value. For the sake of simplicity, it is assumed that $\gamma_\mathbf{p} = 0$.

\subsection{Resolution}The resolution criterion of $l_k$ under the camera with extrinsic parameter $\mathbf{c}_\mathbf{p}$ is defined as
\begin{equation}
\label{C_k^R}
C^R(\mathbf{L}_k,\mathbf{c}_\mathbf{p}) = \frac{fd_s}{(d_s-f)z^c_k\max(s_u,s_v)},
\end{equation}
where $f$, $d_s$, $s_u$ and $s_v$ are intrinsic parameters of the camera that explained in Table I, $z_k^c$ is the coordinate of $l_k$ on $\mathbf{Z}^c$-axis in $\mathcal{F}^c$.

\subsection{Field of View}
The camera model contains horizontal FOV angles $\varphi_l, \varphi_r$ and vertical FOV angles $\varphi_t, \varphi_b$. For landmark $l_k$, if the geometrical constraints
\begin{equation*}
\label{fov1}
z_k^c > 0,
\end{equation*}
\begin{equation*}
\label{fov2}
\frac{\sqrt{{x_k^c}^2+{y_k^c}^2}}{z_k^c} \leq \tan{(\min{(\varphi_l, \varphi_r, \varphi_t, \varphi_b)})},
\end{equation*}
are satisfied, then it is said that $l_k$ falls in the FOV of the camera. The FOV criterion of $l_k$ under the camera with extrinsic parameter $\mathbf{c}_\mathbf{p}$ is defined as
\begin{equation}
\label{C_k^FOV}
C^{FOV}(\mathbf{L}_k,\mathbf{c}_\mathbf{p}) =
\begin{cases}
1 & \text{fall in FOV} \\
0 & \text{otherwise}.
\end{cases}
\end{equation}

\subsection{Focus}
For landmark $l_k$, given diameter $\delta \in \mathbb{R}^+$ of the permissible circle of confusion in pixel, if $z_k^c$ falls in the range $d_n \le z_k^c \le d_f$, then it is said that $l_k$ is focused by the camera, where
\begin{equation*}
\label{d_n_d_f}
\vspace{1pt}
\begin{aligned}
d_n =&~ \frac{d_a d_s f}{d_a f + \delta \min(s_u,s_v)(d_s-f)},\\[1pt]
d_f =&~ \frac{d_a d_s f}{d_a f - \delta \min(s_u,s_v)(d_s-f)}\\[1pt]
\end{aligned}
\end{equation*}
are the near-field depth and far-field depth, respectively, with $d_n < d_f$. The focus criterion of $l_k$ under the camera with extrinsic parameter $\mathbf{c}_\mathbf{p}$ is defined as
\begin{equation}
\label{C_k^F}
C^F(\mathbf{L}_k,\mathbf{c}_\mathbf{p}) =
\begin{cases}
1 & \text{focused} \\
0 & \text{otherwise}.
\end{cases}
\end{equation}

\subsection{Occlusion}
Assume that for the $k^{th}$ landmark, it has a virtual diameter $\nu_k \in \mathbb{R}^+$. Let $\zeta_k^c\in[0,\pi]$ be the elevation angle between $l_k$ and the camera, which is the angle between vector $\mathbf{n}_k$ and vector $\mathbf{s}_k\mathbf{O}^c$, then
\begin{equation*}
\label{zeta_k^c}
\zeta_k^c = \arccos{\frac{\mathbf{n}_k^\mathrm{T}(\boldsymbol{\varsigma}-\mathbf{s}_k)}{\|\boldsymbol{\varsigma}-\mathbf{s}_k\|}}.
\end{equation*}

The landmark $l_k$ is considered occluded in the camera with extrinsic parameter $\mathbf{c}_\mathbf{p}$ if any of the following conditions are met:
\begin{equation}
\label{occlusion1}
\zeta_k^c\ge\pi/2,
\end{equation}
\begin{equation}
\label{occlusion2}
\|\mathbf{s}_j-\mathbf{p}\|\sqrt{1-\left(\frac{(\mathbf{s}_j-\mathbf{p})^\mathrm{T}(\mathbf{s}_k-\mathbf{p})}{\|\mathbf{s}_j-\mathbf{p}\|\|\mathbf{s}_k-\mathbf{p}\|}\right)^2} \le \nu_k,
\end{equation}
where $\mathbf{s}_j$ represents the position of any other landmark except $l_k$. Equation (\ref{occlusion1}) means the camera cannot see the front face of the landmark, while Equation (\ref{occlusion2}) indicates $l_k$ is occluded in other landmarks. Then the occlusion criterion of $l_k$ under the camera with extrinsic parameter $\mathbf{c}_\mathbf{p}$ is defined as
\begin{equation}
\label{C_k^O}
C^O(\mathbf{L}_k,\mathbf{c}_\mathbf{p}) =
\begin{cases}
0 & \text{occluded} \\
1 & \text{otherwise}.
\end{cases}\vspace{-6pt}
\end{equation}

\subsection{Multiple Coverage Probability}

As the camera is moving, its position and orientation are changing over time. The captured landmarks information is also time-varying. Assume that the position of the camera is a random point in the 3-D reachable area, the concept of $n$-ple coverage probability is proposed to evaluate the chance of this camera to cover no less than $n$ landmarks with the consideration of its geometry properties.   


The spherical cap $S$ is defined as a set of continuous and/or discontinuous partial areas on the unit sphere. Suppose a camera is placed at $\mathbf{p}$, the optical axis $\mathbf{Z}^c$ of the camera can point to any direction. Taking $\mathbf{p}$ as the starting point, the end points of all possible unit vectors along the optical axis will form a unit sphere $S^{RU\!S}$ called reachable unit sphere. The reachable unit sphere contains all possible orientations of the camera at a fixed position.

However, not all camera orientations meet the requirements of the visual task. Following the idea in \cite{lei2021radial}, the coverage strength of $l_k$ under the camera with extrinsic parameter $\mathbf{c}_\mathbf{p}$ is introduced as
\begin{align}
\label{Cs_k}
Cs&(\mathbf{L}_k,\mathbf{c}_\mathbf{p}) = \nonumber \\
&C^R(\mathbf{L}_k,\mathbf{c}_\mathbf{p})C^{FOV}\hspace{-3pt}(\mathbf{L}_k,\mathbf{c}_\mathbf{p})C^F(\mathbf{L}_k,\mathbf{c}_\mathbf{p})C^O(\mathbf{L}_k,\mathbf{c}_\mathbf{p}),
\end{align}
which leads to the concept of the coverage spherical cap.
\begin{definition}
Given a coverage strength threshold $thold\in\mathbb{R}^+$, there exists a set of region $S^C_k(\mathbf p)$ on the reachable unit sphere, such that when the optical axis $\mathbf{Z}^c$ passes through this region, $Cs(\mathbf{L}_k,\mathbf{c}_\mathbf{p}) \geq thold$ is satisfied. Then $S^C_k(\mathbf p)$ is called the coverage spherical cap of $l_k$ at $\mathbf{p}$.
\end{definition}

In the sequel, we define the $n$-ple coverage spherical cap, indicating at least $n$ landmarks can be covered at the position $\mathbf{p}$. 
\begin{definition}
Let $Num(S^C_k(\mathbf p))\in\mathbb{N}$ be the number of coverage spherical caps that the optical axis $\mathbf{Z}^c$ passes through. Given a task parameter $n\in\mathbb{N}$ called the qualified threshold of landmarks, there exists a set of region $S^n(\mathbf{p})$ on the reachable unit sphere, such that when the optical axis $\mathbf{Z}^c$ passes through this region, $Num(S^C_k(\mathbf p)) \geq n$ is satisfied. Then $S^n(\mathbf{p})$ is called the $n$-ple coverage spherical cap at $\mathbf{p}$.
\end{definition}

\begin{figure}[!t]
\centering
\includegraphics[scale=1]{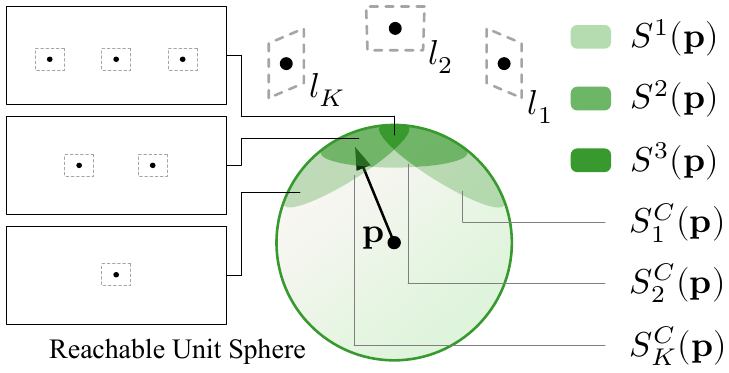}
\caption{An example of multiple coverage spherical cap.}
\label{multiple_coverage_spherical_cap}
\end{figure}

An example is shown in Fig. \ref{multiple_coverage_spherical_cap} to illustrate the idea of $n$-ple coverage spherical cap. For each landmark, a coverage spherical cap can be found on the reachable unit sphere at $\mathbf{p}$. For example, when the optical axis of a camera $\mathbf{p}$ passes through $S^C_1(\mathbf p)$, it indicates that the camera can cover $l_1$. The optical axis of the camera is presented by the black arrow and passes through $2$-ple coverage spherical cap. That is, the optical axis of the camera passes through at least two coverage spherical cap. This shows that the camera can cover at least two landmarks at the same time. The value of $n$ is usually task dependent. The coverage spherical cap at $\mathbf{p}$ may be an empty set for some landmark setting.

In order to characterize possible orientations of the camera, the orientation probability density is introduced. Let $D(\alpha_\mathbf{p},\beta_\mathbf{p}) \in [0,1]$ be the orientation probability density function (PDF) of the camera placed at $\mathbf{p}$, the following conditions are satisfied:
\begin{equation*}
\label{D1}
D(\alpha_\mathbf{p},\beta_\mathbf{p}) \geq 0,
\end{equation*}
\begin{equation*}
\label{D2}
\iint_{S^{RU\!S}}\!\!D(\alpha_\mathbf{p},\beta_\mathbf{p})\rm{d}\beta_\mathbf{p}\rm{d}\alpha_\mathbf{p} = 1.
\end{equation*}
This is a very general form of PDF. In our simulations  the uniform distribution is used while in experiment, this PDF is estimated from data collected, which is also the uniform distribution. Other form of PDFs can be used as well.  Finally, the $n$-ple coverage probability at $\mathbf{p}$ can be formulated as
\begin{equation}
\label{P_n}
P_n(\mathbf{p}) = \iint_{S^n(\mathbf{p})}\!\!D(\alpha_\mathbf{p},\beta_\mathbf{p})\rm{d}\beta_\mathbf{p}\rm{d}\alpha_\mathbf{p},
\end{equation}
which represents the probability that the camera can cover at least $n$ landmarks at the same time when the camera is placed arbitrarily at $\mathbf{p}$. This multiple coverage probability is used as the basis of landmark deployment optimization.

\section{Landmark Deployment Optimization}
\label{section4} 
With the characterization of multiple coverage probability at each position $\mathbf{p}$ in a given 3-D reachable area of $\Omega$, landmark deployment can be formulated as an optimization problem, maximizing the chance of at least $n$ landmarks being covered by the camera at each point. This becomes a global optimization problem for a given 3-D space by exploring every point in it. A very natural selection to solve this global optimization is standard genetic algorithms (SGAs) 
\cite{chen2004automatic}, though other techniques such as particle swarm optimization (PSO) \cite{morsly2011particle} can be also used. It is noted that SGAs cannot guarantee the monotonic convergence \cite{lei2021radial}. In order to speed up the convergence by using monotonic convergence, 
a modified genetic algorithm called elimination genetic algorithm (EGA) is proposed to find this optimum.

\subsection{Cost Function}

First of all, a coverage probability threshold $thold_p\in[0,1]$ is given. If the $n$-ple coverage probability of $\mathbf{p}$ satisfies $P_n(\mathbf{p}) \geq thold_p$, then $\mathbf{p}$ is said to be qualified.
The selection of this threshold is usually case-dependent. In simulations, we show  how the choices of $(n,thold_p)$ will affect landmark deployment. Intuitively, the larger $n$ provides more robustness in finishing the visual task such as the visual observer. Usually, a large $n$ also requires a smaller $thold_p$ as demonstrated in simulations in Section V.A.

The cost function thus can be formulated as
\begin{equation}
\label{H}
\mathcal{H}(\mathbf{L}_1,\mathbf{L}_2,\!\cdots\!,\!\mathbf{L}_K\!,\mathbf{p}) = \int_\Omega\!Qual(\mathbf{L}_1,\mathbf{L}_2,\!\cdots\!,\!\mathbf{L}_K\!,\mathbf{p})Rel(\mathbf{p})\rm{d}\mathbf{p},
\end{equation}
which represents the total volume of the qualified area, where
\begin{equation}
\label{Qual}
Qual(\mathbf{L}_1,\mathbf{L}_2,\!\cdots\!,\!\mathbf{L}_K\!,\mathbf{p}) =
\begin{cases}
1 & \text{qualified} \\
0 & \text{otherwise}.
\end{cases}
\end{equation}
Here $Rel(\mathbf{p})$ is the relevance function representing the importance of different positions in the reachable area. Similar to the coverage probability threshold, the choice of it is also case-dependent. In our simulations and experiments, this function is uniformly distributed over the space of interests. In some applications, if some sub-region of the space is more important than other regions, this function can be tuned accordingly.

Finally, we obtain the following optimization problem for the directional landmark deployment:
\begin{equation}
\label{argmaxH}
\mathop{\arg\max}_{\mathbf{L}_1,\cdots,\mathbf{L}_K}\mathcal{H}(\mathbf{L}_1,\mathbf{L}_2,\!\cdots\!,\!\mathbf{L}_K\!,\mathbf{p}),~~~
s.t. ~\mathbf{p}\!\in\!\Omega.
\end{equation}

\subsection{Elimination Genetic Algorithm}

Genetic algorithm is a good candidate to optimize the cost function. However SGA algorithms cannot guarantee the monotonic of convergence in terms of the cost function, resulting a slower convergence. Hence, fast convergence of global optimization is preferred. This motivates new elimination operation for SGA, which is called EGA, speed up the convergence and promote the random search needed for the cost function. It is worth noting that EGA is not a global optimal algorithm, and the cost function constructed in this paper is also difficult to optimize to the global optimal. The purpose of the optimization is to improve the performance of landmark deployment as much as possible.

\begin{figure}[!t]
\centering
\includegraphics[scale=1]{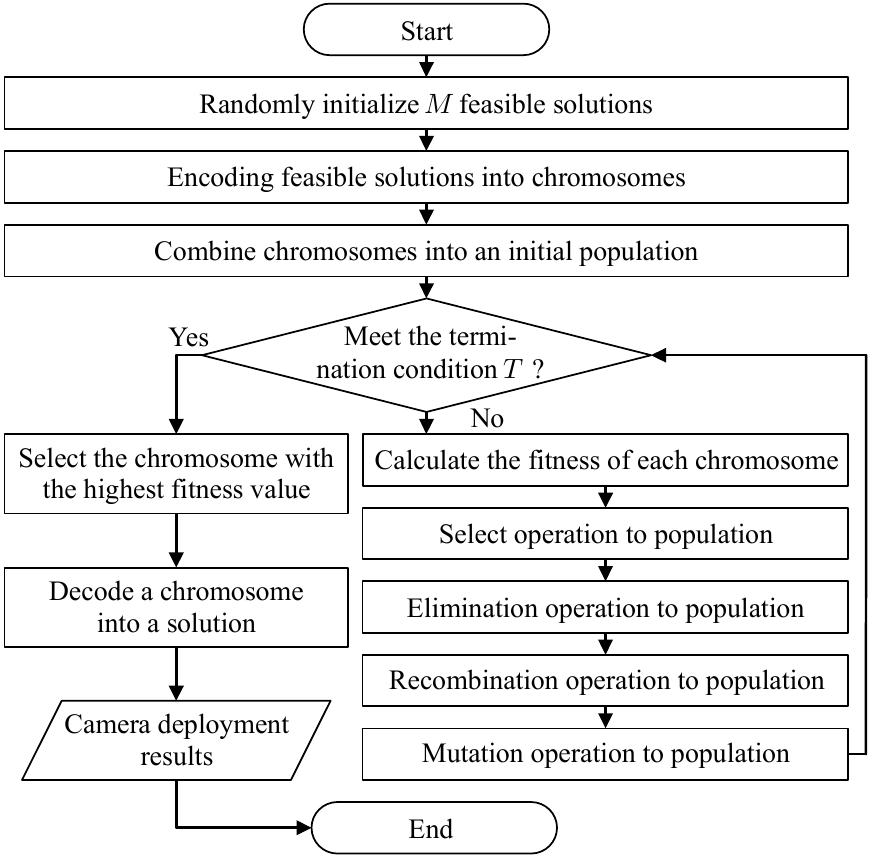}
\caption{Flowchart of elimination genetic algorithm.}
\label{EGA_flowchart}
\end{figure}

The EGA follows the similar steps of SGA in generating population and fitness function. Each feasible solution is encoded as a string called a chromosome, and several chromosomes form a population. To solve landmark deployment problem in (\ref{argmaxH}), each feasible solution is reconstructed into a chromosome, this process is called encoding.
 Suppose $\{\mathbf{L}_1,\mathbf{L}_2,\cdots,\mathbf{L}_K\}$ is a feasible solution, then the encoded chromosome becomes
$
\mathbf{CH} = \{x_1,y_1,z_1,\rho_1,\eta_1,\mu_1,\!\cdots\!,\!x_K,y_K,z_K,\rho_K,\eta_K,\mu_K\}
$.
$M\in\mathbb{N}^+$ chromosomes are formed into a population. Let 
$
\eth = \{\mathbf{CH}_1,\mathbf{CH}_2,\cdots,\mathbf{CH}_M\}$ be a population, where $\mathbf{CH}_m$ represents the $m^{th}
$ chromosome ($m=1,2,\cdots, M$), and the $t^{th}$ population is represented by $\eth_t$. The fitness function is defined as the same as the cost function:
$
\mathrm{Fit}(\mathbf{CH}) = \mathcal{H}(\mathbf{L}_1,\mathbf{L}_2,\!\cdots\!,\!\mathbf{L}_K\!,\mathbf{p}).
$

\begin{figure}[!t]
\centering
\includegraphics[scale=1]{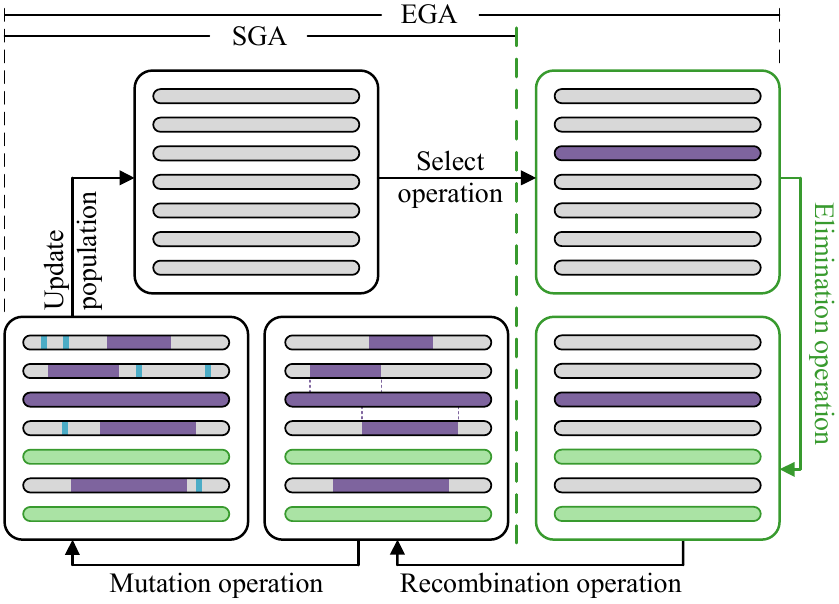}
\caption{Difference between SGA and EGA}
\label{population_update}
\end{figure}

The difference of EGA and SGA is shown in Fig. \ref{population_update}. It is noted that in order to achieve monotonic convergence, the existing optimal selection operations such as recombination operation and mutation operation proposed in \cite{lei2021radial} are used in EGA. The parameters of the EGA are summarized in Table \ref{parameters_of_EGA}. Moreover, a new elimination operation are added in EGA, in which $Q$ Chromosomes in the population with the worst fitness are eliminated and replaced by newly generated chromosomes. The chromosomes other than the survival chromosome and the newly generated chromosomes constitute the recombinant pool.

It is noting that the parameters satisfy $Q+1 \leq M$ and $\Upsilon_{max} \le L$. In Fig. \ref{population_update}, the gray bar represents the chromosome, the purple bar represents the survival chromosome, and the green bar represents the mutated gene. After reaching the termination condition, it is necessary to decode the chromosome with the highest fitness in the population to obtain the final landmark deployment solution. The decoding process is the inverse of the encoding process. The parameters of the elimination genetic algorithm can be included in a nine-tuple:
\begin{equation*}
\label{EGA}
EG\hspace{-1pt}A = (W,\mathrm{Fit},\eth_0,M,Q,\Upsilon_{min},\Upsilon_{max},\Psi,T),
\end{equation*}
where $W$ is the coding scheme of the chromosome and $\eth_0$ is the initial population.

The flowchart of the elimination genetic algorithm is described in Fig. \ref{EGA_flowchart}. This leads to the updated operations of the population as shown in Fig. \ref{population_update}.

\begin{table}[!t]
\renewcommand{\arraystretch}{1.3}
\caption{Parameters of Elimination Genetic Algorithm}
\label{parameters_of_EGA}
\centering
\begin{tabular}{ll}
\hline\hline
\specialrule{0em}{0pt}{2pt}
Parameter & Description \\
\specialrule{0em}{2pt}{0pt}
\hline
\specialrule{0em}{2pt}{0pt}
$M\in\mathbb{N}^+$ & Size of population \\
$L\in\mathbb{N}^+$ & Length of chromosome (number of genes) \\
$Q\in\mathbb{N}^+$ & Number of eliminated chromosome \\
$\Upsilon_{min},\Upsilon_{max}\in\mathbb{N}^+$ & Minimum / Maximum length of recombination \\
$\Psi\in(0,1]$ & Mutation probability of each gene \\
$T$ & Algorithm termination condition \\
\specialrule{0em}{2pt}{0pt}
\hline\hline
\end{tabular}
\end{table}

\section{Simulation and Experiment}
\label{section5}
In this section, simulations and experiments are designed to validate the capability of the proposed landmark deployment method and its impact on the visual observer design discussed in the motivating example. 

\subsection{Simulation Results of Landmark Deployment}

We first construct indoor landmark deployment scenario.
Suppose that the indoor space is $750 \times 500 \times 600$ cm in length, width and height, respectively. 
The landmarks are distributed on the six walls of it. The number of landmarks is set to be $K = 90$. For each landmark, it has five degrees of freedom: three in position, and two in orientation which is the yaw angle and the pitch angle. 
The reachable area of the camera sensor in the space is also a cuboid, the geometric center of which coincides with the center of the indoor space, with the space $\Omega$ of reachable area being $600 \times 350 \times 450$ cm in length, width and height, respectively. The reachable area is uniformly discretized into $1040$ grids. The yaw angle and pitch angle of the camera are discretized as $\pi/12$ ($\alpha_\mathbf{p} = -\pi,-\frac{11}{12}\pi, \cdots, \frac{11}{12}\pi, \beta_\mathbf{p} = -\frac{11}{24}\pi,-\frac{9}{24}\pi, \cdots, \frac{11}{24}\pi$). That is, the camera has $24\times12=288$ grids on each reachable unit sphere. Let $\Omega_d$ denote the set of all discrete 3-D coordinates in the reachable area, then the discrete cost function we optimized in the simulation can be rewritten as
\begin{equation*}
\label{H2}
\mathcal{H}(\mathbf{L}_1,\mathbf{L}_2,\!\cdots\!,\!\mathbf{L}_K\!,\mathbf{p}) = \sum_{\mathbf{p}\in\Omega_d}\!Qual(\mathbf{L}_1,\mathbf{L}_2,\!\cdots\!,\!\mathbf{L}_K\!,\mathbf{p})Rel(\mathbf{p}).
\end{equation*}
The intrinsic parameters of the camera in the visual task are shown in Table \ref{intrinsic_parameters_in_simulation}. In the simulation, the diameter of permissible circle of confusion is $\delta = 4$ pixel. Next several comparisons will be performed.

\begin{table}[!t]
\renewcommand{\arraystretch}{1.3}
\caption{Intrinsic Parameters of the Camera in Simulation}
\label{intrinsic_parameters_in_simulation}
\centering
\begin{tabular}{ll}
\hline\hline
\specialrule{0em}{0pt}{2pt}
Description & Parameter \& Value \\
\specialrule{0em}{2pt}{0pt}
\hline
\specialrule{0em}{2pt}{0pt}
Lens focal length & $f = 5$ mm \\
Horizontal pixel dimensions & $s_u = 0.0058$ mm/pixel \\
Vertical pixel dimensions & $s_v = 0.0058$ mm/pixel \\
Principle point &
$
\mathbf{o} = [\begin{IEEEeqnarraybox*}[][c]{,c/c,}
800 & 600
\end{IEEEeqnarraybox*}]^\mathrm{T}
$ in pixel \\
Image width & $w = 1600$ pixel \\
Image height & $h = 1200$ pixel \\
Effective aperture diameter of optical lens & $d_a = 10$ mm \\
Focusing distance & $d_s = 1778$ mm \\
\specialrule{0em}{2pt}{0pt}
\hline\hline
\end{tabular}
\end{table}


For simplicity, the permissible number of landmarks is set to $n = 2$. The orientation probability density function of the camera is defined as $D(\alpha_\mathbf{p},\beta_\mathbf{p}) = 1/288$, that is, the orientation of the camera satisfies a uniform distribution. In later experiment, $D(\alpha_\mathbf{p},\beta_\mathbf{p})$ is estimated from the experimental data. Other application specific distributions can be considered as well. The coverage probability threshold is $thold_p = 0.65$. With this setting of parameters, we can compare the performance of SGA and EGA.

The parameters of both SGA and EGA are set as follows: The size of the population is $M=30$, the length of the chromosome is $L=5\times K=450$ due to each camera has five degrees of freedom. In EGA, the number of eliminated chromosome is set to $Q=7$, and the minimum and maximum length of recombination are $\Upsilon_{min}=100,\Upsilon_{max}=300$, respectively, and the mutation probability of each gene is $\Psi=0.1$. Moreover, to be fair for comparison of different algorithms, all the optimization algorithms in this subsection are set to terminate at the $400^{th}$ iteration.

The following performance indices are employed to evaluate the performance the landmark deployment:
\begin{enumerate}[\IEEEsetlabelwidth{12)}]
\item
\emph{Qualified Ratio:} The volume or the number proportion of the qualified coordinates to all coordinates in the reachable area.
\item
\emph{Average Coverage Probability:} The average of the $n$-ple coverage probability of all coordinates in the reachable area.
\item
\emph{Maximum Coverage Probability:} The maximum $n$-ple coverage probability of all coordinates in the reachable area.
\end{enumerate}

\begin{figure*}[!t]
\centering
\subfloat[The initial random landmark deployment]{\includegraphics[scale=1]{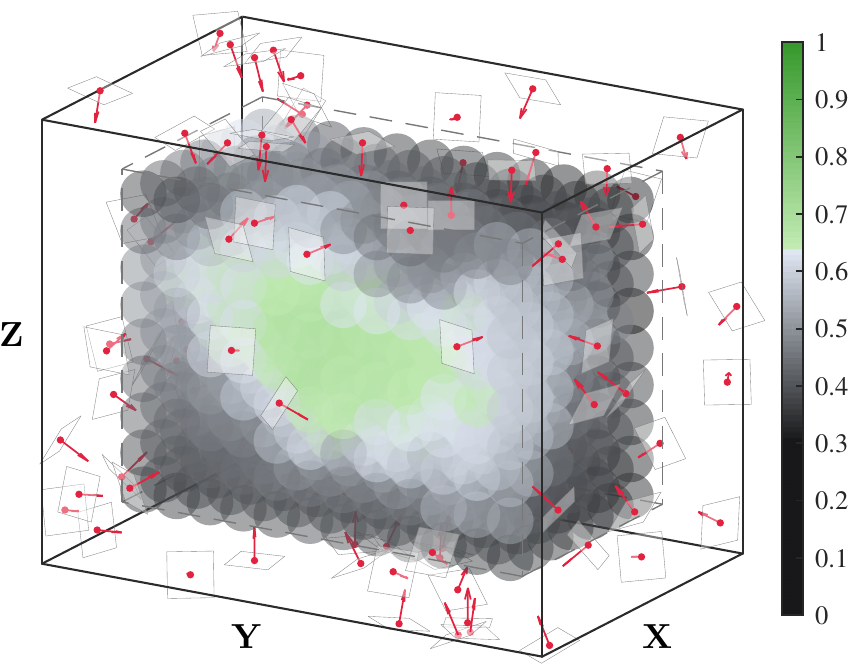}
\label{simulation_results_initial_EGA}}
\hfil
\subfloat[The final landmark deployment using EGA]{\includegraphics[scale=1]{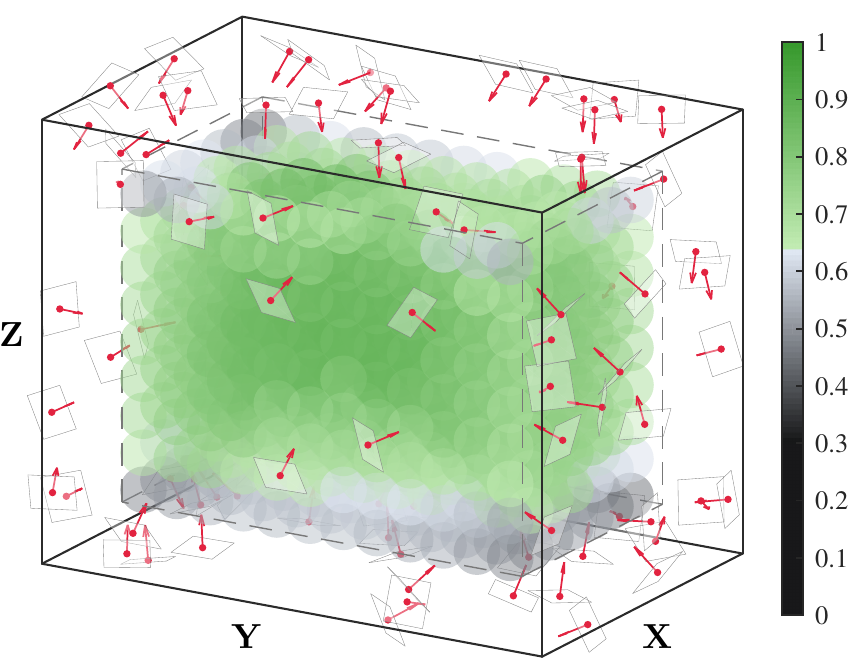}
\label{simulation_results_final_EGA}}
\hfil
\subfloat[The final landmark deployment using SGA]{\includegraphics[scale=1]{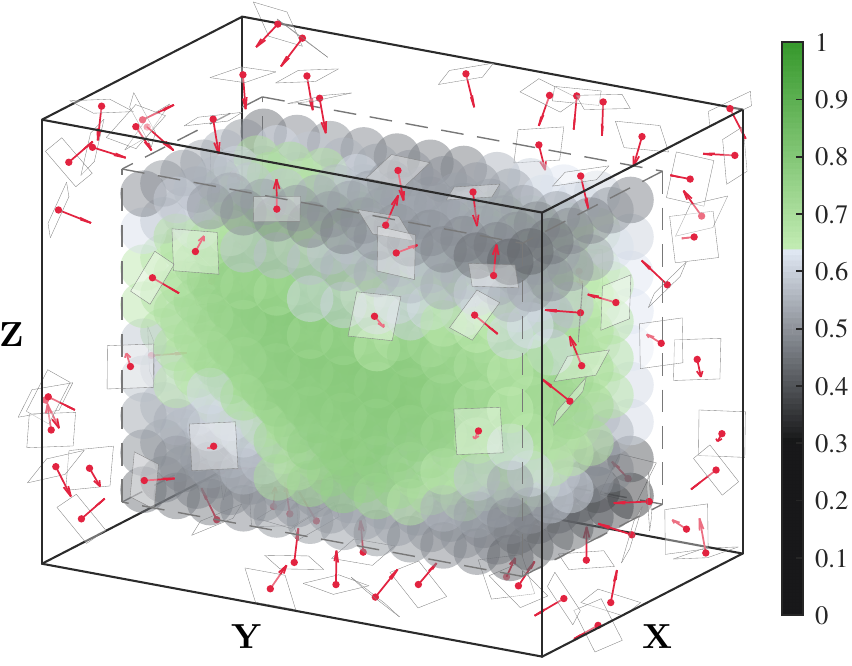}
\label{simulation_results_final_GA}}
\hfil
\subfloat[The uniform landmark deployment]{\includegraphics[scale=1]{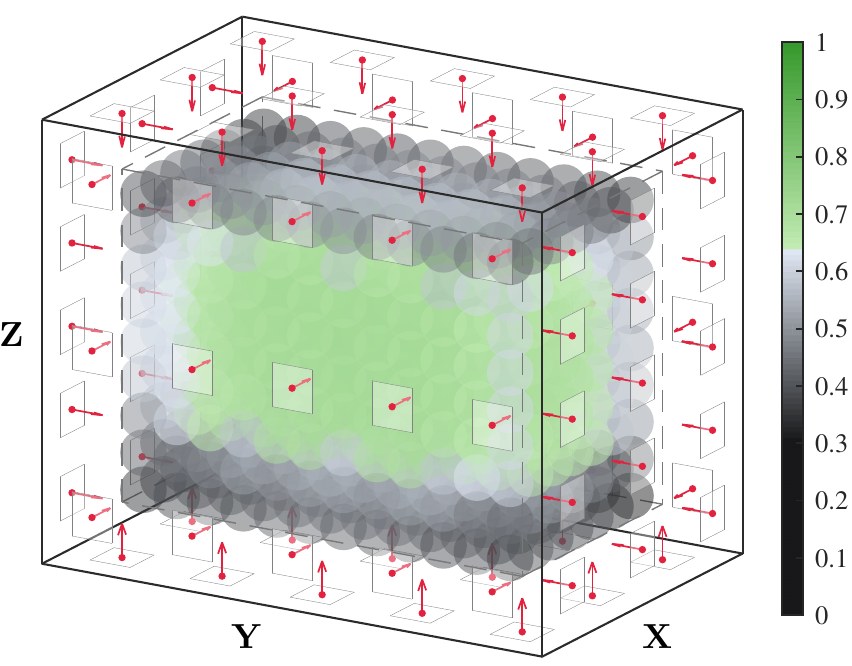}
\label{simulation_results_UNI}}
\caption{Simulation results of the landmark deployment.}
\label{simulation_results}
\end{figure*}

Next two figures further compare the performance of SGA and EGA. Fig. \ref{simulation_results} visually shows the difference between SGA and EGA. In each sub-figure, the black edge cuboid represents the indoor space where the landmarks are explored. The gray dotted edge cuboid represents the reachable area of the camera. The red arrows indicates the position and orientation of the landmarks. Colored balls are distributed in the reachable area evenly, each ball represents a discrete 3-D coordinate, and the color represents the value of the $2$-ple coverage probability. The green ball means that the coverage probability is qualified at this position, while the gray ball means that the coverage probability is unqualified at this position. Fig. \ref{simulation_results_initial_EGA} is a random initialization. It shows that the unqualified area occupies most of the reachable area at the initial setting. After $400$ iterations, the optimal solutions shown in Fig. \ref{simulation_results_final_EGA} with the majority of this areas have become qualified, and the coverage probability of unqualified areas has also increased. The results optimized by SGA are shown in Fig. \ref{simulation_results_final_GA}. This shows that the proposed EGA can improve the optimal performance. 

\begin{figure}[!t]
\centering
\includegraphics[scale=1]{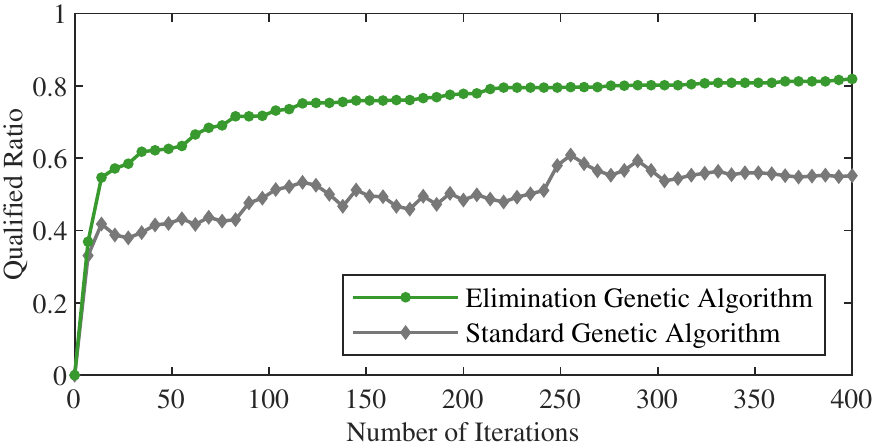}
\caption{Comparison of the qualified ratio curves with different algorithms.}
\label{simulation_curve}
\end{figure}

In addition, a landmark setting used uniform distribution over the space of interest, which is also the most common landmark deployment method in various visual tasks \cite{zhu2014robot}, is shown in Fig. \ref{simulation_results_UNI}. 
$90$ landmarks are evenly deployed on 6 walls of the indoor space with the coverage probability of the qualified areas is computed. It is noted that many areas are still unqualified, showing that the uniform contribution cannot always get good performance.  
Fig. \ref{simulation_curve} shows that EGA indeed converges monotonically and it converges faster than that of SGA. Table \ref{evaluation_metrics} compares the performance between SGA and EGA, showing the effectiveness of EGA.

\begin{table}[!t]
\renewcommand{\arraystretch}{1.3}
\caption{Evaluation Metrics for Different Landmarks Deployments}
\label{evaluation_metrics}
\centering
\begin{tabular}{lccc}
\hline\hline
\specialrule{0em}{0pt}{2pt}
Deployment & \hspace{3pt}\makecell{Qualified \\Ratio}\hspace{3pt} & \hspace{3pt}\makecell{Average \\Coverage \\Probability}\hspace{3pt} & \makecell{Maximum \\Coverage \\Probability} \\
\specialrule{0em}{2pt}{0pt}
\hline
\specialrule{0em}{0pt}{2pt}
Final Deployment (EGA) & $81.88\,\%$ & $75.00\,\%$ & $91.69\,\%$ \\
Final Deployment (SGA) & $53.44\,\%$ & $64.45\,\%$ & $84.31\,\%$ \\
Uniform Deployment & $43.92\,\%$ & $60.16\,\%$ & $74.46\,\%$ \\
Initial Deployment & $11.11\,\%$ & $52.31\,\%$ & $71.08\,\%$ \\
\specialrule{0em}{2pt}{0pt}
\hline\hline
\end{tabular}
\end{table}

\begin{table}[!t]
\renewcommand{\arraystretch}{1.3}
\caption{The Impact of Parameters on Deployment Performance}
\label{parameters_impact}
\centering
\begin{tabular}{llccc}
\hline\hline
\specialrule{0em}{0pt}{2pt}
\multicolumn{2}{l}{\hspace{3pt}Parameter} & \multirow{2}{*}{\hspace{3pt}\makecell{Qualified \\Ratio}\hspace{3pt}} & \multirow{2}{*}{\hspace{3pt}\makecell{Average \\Coverage \\Probability}\hspace{3pt}} & \multirow{2}{*}{\makecell{Maximum \\Coverage \\Probability}} \\
\hspace{3pt}$n$\hspace{10pt} & $thold_p$ & & & \\
\specialrule{0em}{5pt}{0pt}
\hline
\specialrule{0em}{0pt}{2pt}
\hspace{3pt}$2$ & $0.50$ & $97.75\,\%$ & $75.00\,\%$ & $91.69\,\%$ \\
\hspace{3pt}$2$ & $0.65$ & $81.88\,\%$ & $75.00\,\%$ & $91.69\,\%$ \\
\hspace{3pt}$2$ & $0.80$ & $39.68\,\%$ & $75.00\,\%$ & $91.69\,\%$ \\
\hspace{3pt}$3$ & $0.50$ & $73.94\,\%$ & $59.70\,\%$ & $83.08\,\%$ \\
\hspace{3pt}$3$ & $0.65$ & $37.04\,\%$ & $59.70\,\%$ & $83.08\,\%$ \\
\hspace{3pt}$3$ & $0.80$ & $3.97\,\%$ & $59.70\,\%$ & $83.08\,\%$ \\
\hspace{3pt}$4$ & $0.50$ & $35.85\,\%$ & $46.76\,\%$ & $73.85\,\%$ \\
\hspace{3pt}$4$ & $0.65$ & $7.41\,\%$ & $46.76\,\%$ & $73.85\,\%$ \\
\hspace{3pt}$4$ & $0.80$ & $0.00\,\%$ & $46.76\,\%$ & $73.85\,\%$ \\
\specialrule{0em}{2pt}{0pt}
\hline\hline
\end{tabular}
\end{table}

Besides, in the proposed cost function, the performance of landmark deployment depends on the choice of parameters $(n, thold_p)$. Then, how different choices of them will affect the performance are shown in Table \ref{parameters_impact}. This result suggests that when a larger $n$ is selected, the $thold_p$ needs to be selected smaller to achieve a reasonable qualified ratio.

\subsection{Simulation Results of Visual Observer on SE\emph{(3)} }
This subsection compares the performance of the visual observer on \emph{SE}(3) (see (\ref{X_hat_dot})) by using three landmark deployment settings: the optimized deployment, uniform deployment, and random deployment. To test the landmark information that the camera can capture in various positions and orientations, the camera is set to perform random movement in the reachable area for a sufficient time, such that the camera can collect enough data for analysis.

The performance of the visual observer is mainly measured by the error between the pose of the moveable camera and its observer as
$
Er = \|\hat{X}-X\|^2_\mathrm{F}
$.
The environment model, camera model, and other task parameters are the same as the settings in simulations above. In each set of deployments, the camera has the same initial position and orientation, same linear and angular velocities. The initial pose of the observer is also the same. In this simulation, random noise is not considered. 


\begin{figure}[!t]
\centering
\includegraphics[scale=1]{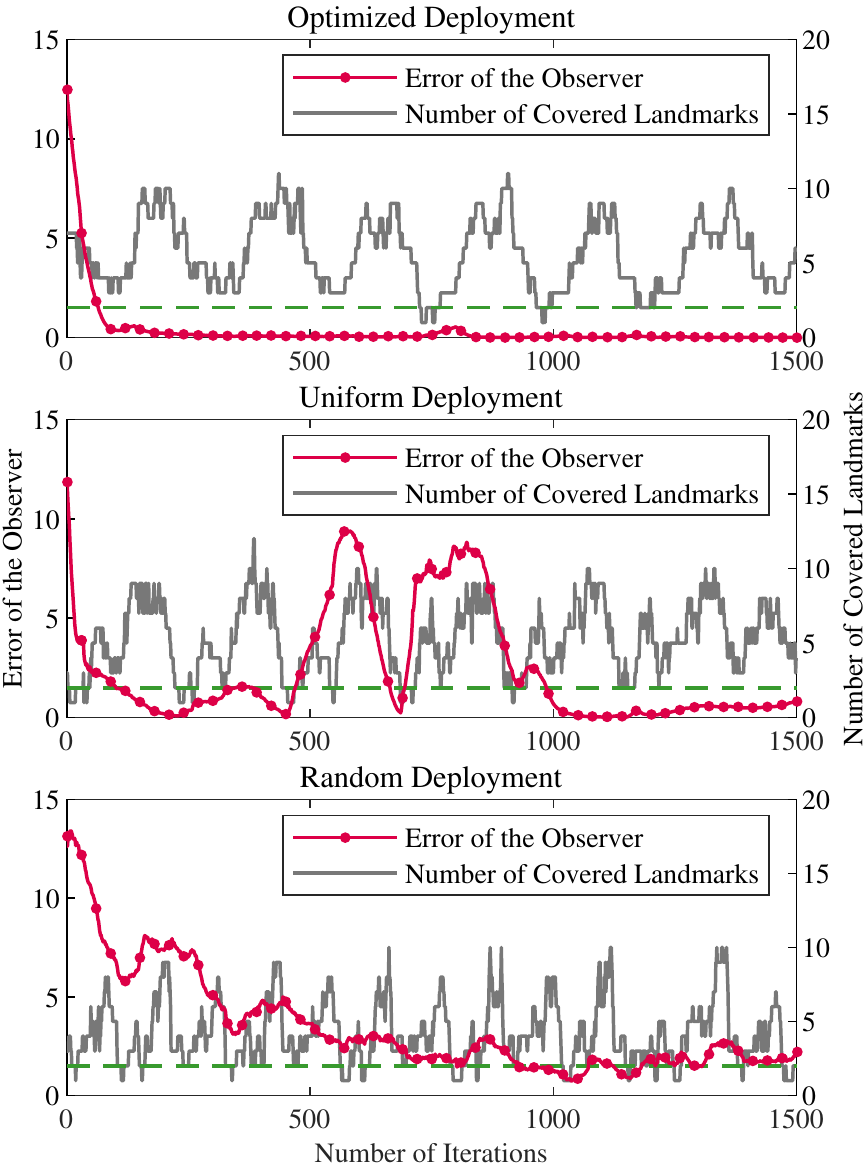}
\caption{Error of different deployment in the observer simulation.}
\label{observer_error}
\end{figure}

The performance of the visual observer on \emph{SE}(3) with different landmark deployments are shown in Fig. \ref{observer_error}. The value of the green dotted line is $2$, indicating $2$-ple coverage probability. The simulation results show that the optimal landmark deployment has the smallest estimation error among three different deployments. 
In the optimized deployment, at $93.40\,\%$ of the time the number of recognized landmarks is qualified, while the $2$-ple coverage qualified ratio calculated in the uniform deployment is $83.87\,\%$, and the coverage qualified ratio calculated in the random deployment is $78.27\,\%$.

%

%

\subsection{Experiment Results of the Visual Observer on SE\emph{(3)}}
\label{section5d}
An experiment is designed to verify the effectiveness of the visual observer on \emph{SE}(3) presented in the motivating example with proposed landmark deployment method. 

The process and results of the experiment are described in detail. In the experiment, a quadcopter UAV, named Pixhawk4 F450 is used to represent the moving robot in (\ref{system state}). An visual observer is implemented using the measurements of the camera on on-board processor of the UAV.

Different from simulations, the orientation probability density $D(\alpha_\mathbf{p},\beta_\mathbf{p})$ needs to be estimated as it varies for different cameras and tripod heads. In this work, an algorithm based on Monte Carlo method \cite{hammersley2013monte} employed to estimate $D(\alpha_\mathbf{p},\beta_\mathbf{p})$. 
This algorithm is briefly summarized as:
\begin{enumerate}
\item Generate random samples by making the movable camera randomly with recorded the yaw angle and pitch angle data for sufficiently long time.
\item Sample the continuous-time collected data at random intervals to obtain random discrete yaw angle and pitch angle measurements.
\item Perform statistical analysis on two sets of measurements to obtain corresponding two discrete probability density function.
\item Check independence of two random variables. 
\item Fit probability density function. 
\end{enumerate}

\begin{figure}[!t]
\centering
\includegraphics[scale=1]{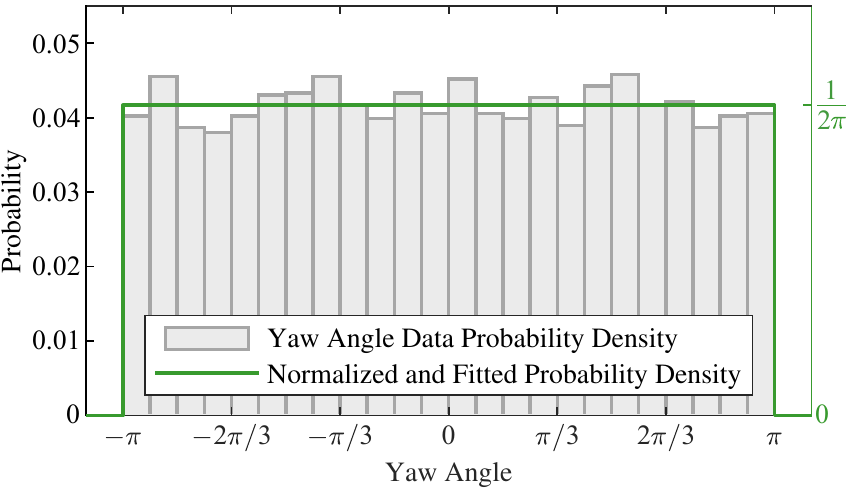}
\caption{Histogram of yaw angle data from experiment.}
\label{PDF_fitting}
\end{figure}

According to this technique, both yaw angle probability density and pitch angle probability density are estimated. 
The gray histogram in Fig. \ref{PDF_fitting} represents the statistical probability density of the captured data sampled at random intervals. The histogram suggests that both of them can be approximated by a uniform distribution. Hence, 
the following normalized orientation probability density of Pixhawk4 F450 is used
\begin{equation}
\label{D_DJI}
D(\alpha_\mathbf{p},\beta_\mathbf{p})\approx D_{PX4}(\alpha)=
\begin{cases}
\frac{1}{2\pi} & -\pi\leq \alpha\leq \pi \\
\hspace{2pt}0 & \text{otherwise}
\end{cases}
\end{equation}
as the green line in Fig. \ref{PDF_fitting}.

Then an indoor platform are created for the landmark deployment experiment, with the space of $320\times 320 \times 170$ cm. The landmarks and the camera installed on this UAV generate the output of the the system (\ref{system state}). The size of the UAV reachable area is $260 \times 260 \times 120$ cm. The landmarks can be deployed on the four walls of the platform, and the number of plate landmarks to be deployed is $K = 80$. The reachable area is uniformly discretized into $147$ coordinates, and the sampling interval of the yaw angle and pitch angle of the camera is $\pi/12$.

\begin{table}[!t]
\renewcommand{\arraystretch}{1.3}
\caption{Intrinsic Parameters of the Camera in Experiment}
\label{intrinsic_parameters_in_experiment}
\centering
\begin{tabular}{ll}
\hline\hline
\specialrule{0em}{0pt}{2pt}
Description & Parameter \& Value \\
\specialrule{0em}{2pt}{0pt}
\hline
\specialrule{0em}{2pt}{0pt}
Lens focal length & $f = 24$ mm \\
Horizontal pixel dimensions & $s_u = 0.0033$ mm/pixel \\
Vertical pixel dimensions & $s_v = 0.0033$ mm/pixel \\
Principle point &
$
\mathbf{o} = [\begin{IEEEeqnarraybox*}[][c]{,c/c,}
960 & 540
\end{IEEEeqnarraybox*}]^\mathrm{T}
$ in pixel \\
Image width & $w = 1920$ pixel \\
Image height & $h = 1080$ pixel \\
Effective aperture diameter of optical lens & $d_a = 8.57$ mm \\
Focusing distance & $d_s = +\infty$ \\
\specialrule{0em}{2pt}{0pt}
\hline\hline
\end{tabular}
\end{table}

\begin{figure}[!t]
\centering
\includegraphics[scale=1]{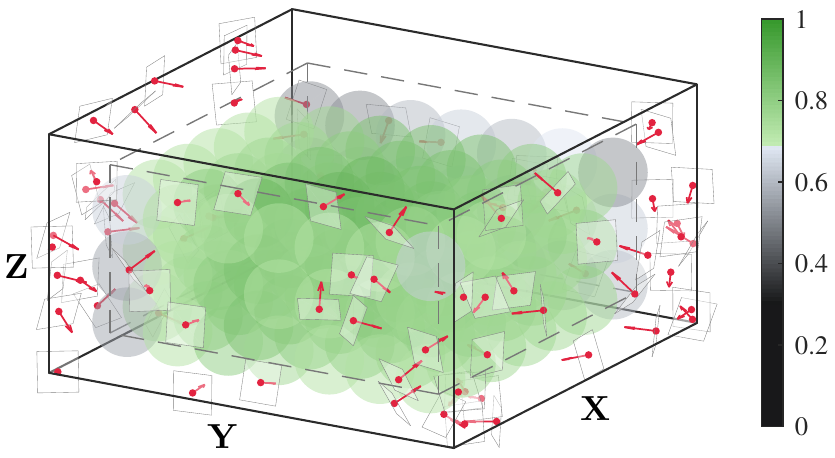}
\caption{Simulation deployment result of indoor platform.}
\label{experiment_simulation_result}
\end{figure}

\begin{figure}[!t]
\centering
\includegraphics[scale=0.16]{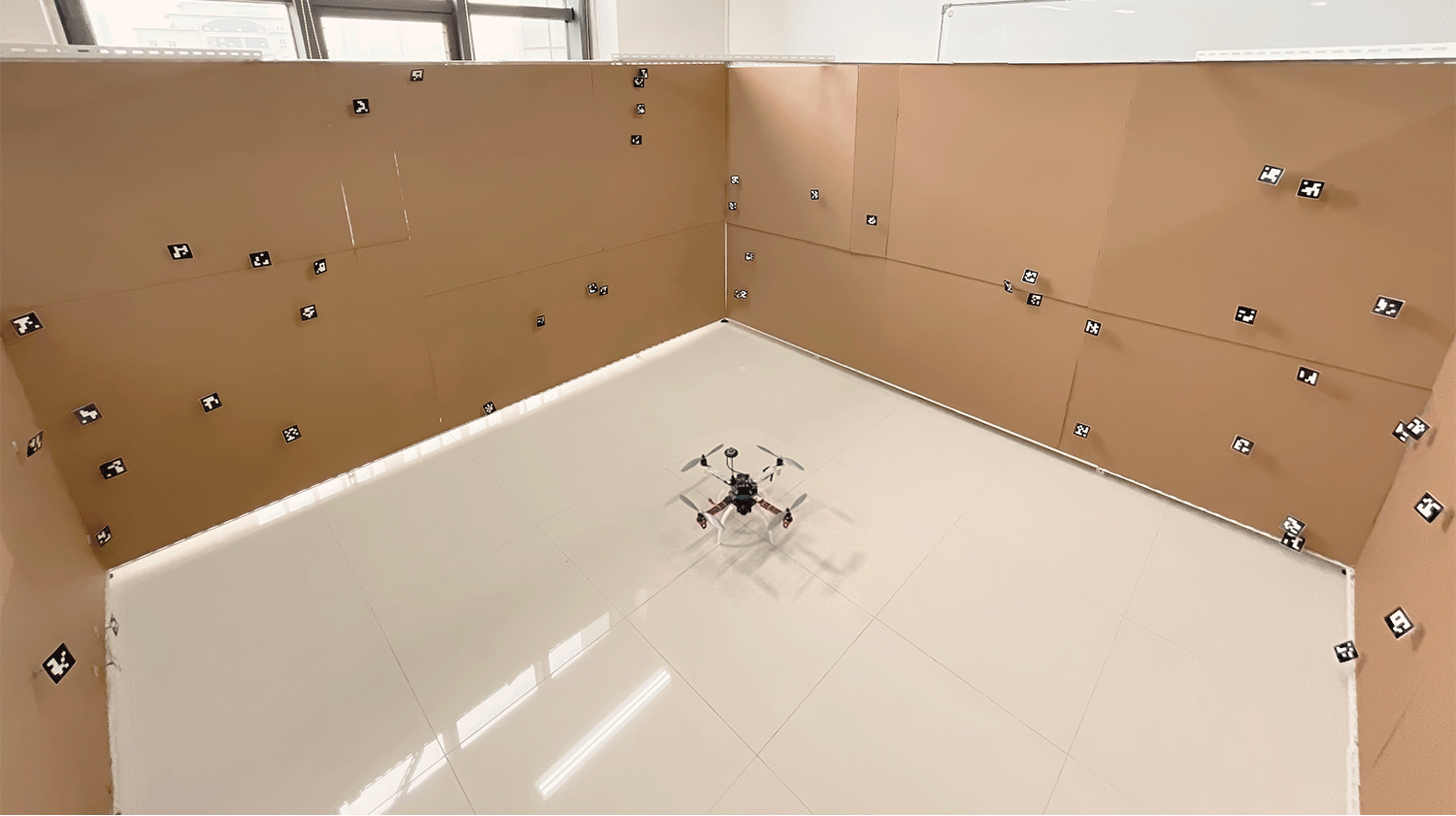}
\caption{Indoor platform and the deployed ArUco markers.}
\label{indoor_platform}
\end{figure}

The intrinsic parameters of the camera on the UAV are listed in Table \ref{intrinsic_parameters_in_experiment}, and the diameter of permissible circle of confusion is $\delta = 4$ pixel. The permissible number of landmarks is set to $n = 2$. As the space of interests in experiment is smaller than one in simulation, the coverage probability threshold is set as $thold_p = 0.7$, which is a bit larger than that in simulation. Based on the experimental setting and estimated normalized orientation probability density of the UAV, the proposed landmark deployment algorithm generates the optimal landmark setting as shown in Fig. \ref{experiment_simulation_result}, Then the ArUco markers are deployed in the indoor platform as shown in Fig. \ref{indoor_platform}.

\begin{figure}[!t]
\centering
\includegraphics[scale=1]{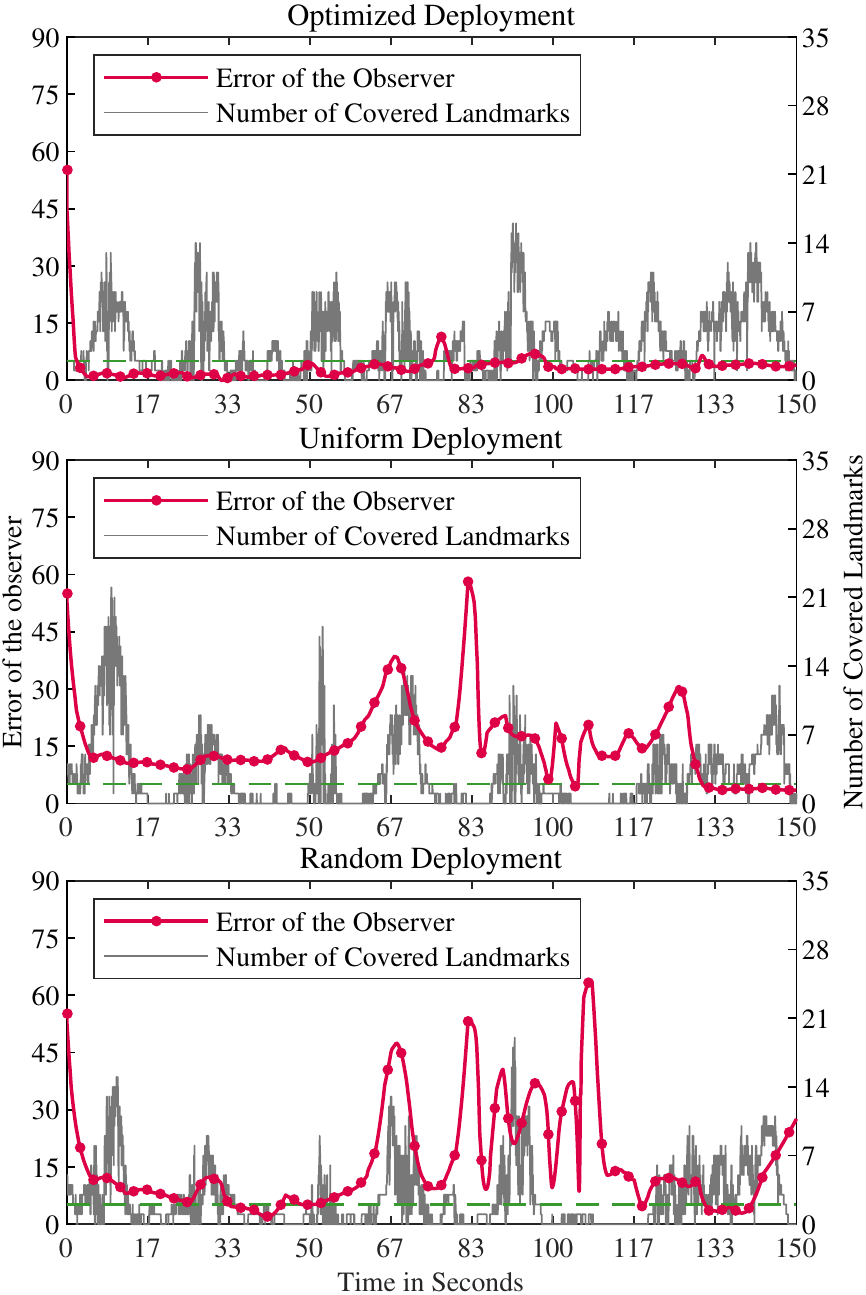}
\caption{Error of different deployment in the observer experiment.}
\label{experiment_result_observer}
\end{figure}

Similar to the simulation in Section V.B, three landmark deployments are tested in the experiment. The UAV is controlled by the computer to fly on the same trajectory  with same linear velocity and angular velocity. The UAV flew for $2.5$ minutes to collect enough experiment data. The number of the captured landmarks are recorded and the observer convergence curve are calculated using the captured information. The performance comparison of the visual observers under $3$ deployments in the experiment is shown in Fig. \ref{experiment_result_observer}.

In the optimized deployment, at $70.14\,\%$ of the time the number of recognized landmarks is qualified, while the $2$-ple coverage qualified ratio calculated in the uniform deployment is $48.90\,\%$, and the coverage qualified ratio calculated in the random deployment is $46.07\,\%$. The experimental results show the effectiveness of the proposed landmark deployment method.

\section{Conclusion}
\label{section6}
In order to improve the performance of the visual observer on \emph{SE}(3) when the directional landmarks and cameras are used to localize the rigid body robotic motion, a new landmark deployment algorithm is proposed in this work. With the consideration of 
geometric models of landmarks and cameras, the novel  criterion called multiple coverage probability, which measures the probability of the landmark being observed by the camera at the fixed position in a 3-D space, is proposed. Consequently, the landmark deployment is formulated as an optimization problem, globally exploring the given space for maximizing the multiple coverage probability. By using the elimination genetic algorithm, the optimal solution is obtained. Simulation and experiment results demonstrate the effectiveness of this landmark deployment technique in improving the performance of the visual observer on \emph{SE}(3).


%



\ifCLASSOPTIONcaptionsoff
  \newpage
\fi



%

\bibliographystyle{IEEEtran}
\bibliography{IEEEabrv,IEEEexample}

%

\vspace*{-2\baselineskip}

\begin{IEEEbiography}[{\includegraphics[width=1in,height=1.25in,clip,keepaspectratio]{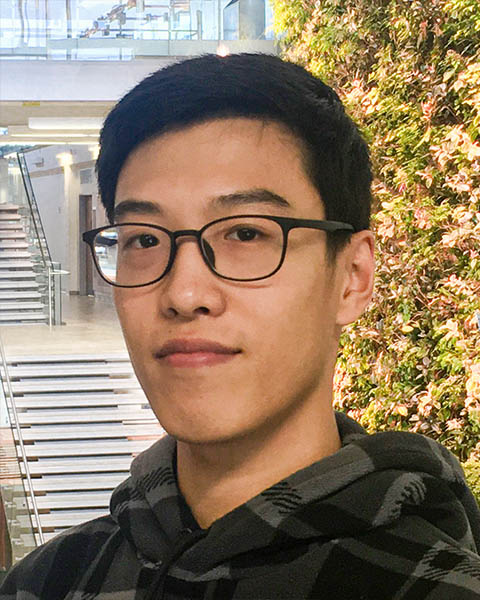}}]{Zike Lei}
received B.Eng. degree in electronic information engineering from Wuhan University of Science and Technology, Hubei, China, in 2017. He is a Ph.D. student in control science and engineering at the School of information Science and Engineering, Wuhan University of Science and Technology. He has been a visiting scholar at University of Windsor, ON, Canada, from September 2019 to August 2020. His research interests include field sensor networks, stereo-camera modeling, and observer-based coverage control.
\end{IEEEbiography}

\vspace*{-2.5\baselineskip}

\begin{IEEEbiography}[{\includegraphics[width=1in,height=1.25in,clip,keepaspectratio]{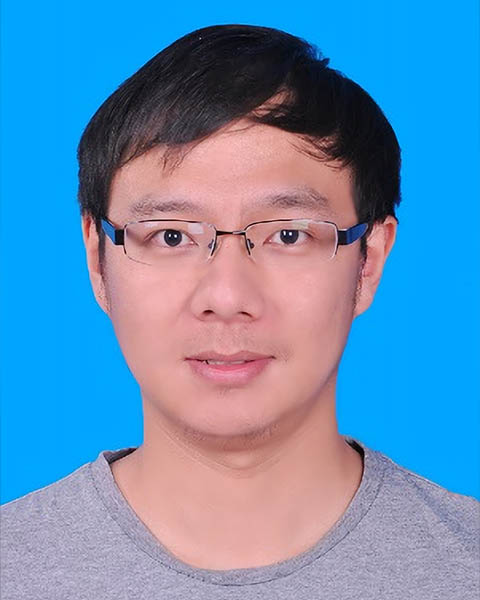}}]{Xi Chen}
received B.S. degree from Huazhong University of Science and Technology, and M.S. degree from Xiamen University, in 2009 and 2012, respectively. He received the Ph.D degree from the University of Newcastle, Australia in 2015. He joined Wuhan University of Science and Technology in 2015, where he is currently an associate professor. His research interests include coverage of sensor network, multi-agent systems and nonlinear system control.
\end{IEEEbiography}

\vspace*{-2.5\baselineskip}

\begin{IEEEbiography}[{\includegraphics[width=1in,height=1.25in,clip,keepaspectratio]{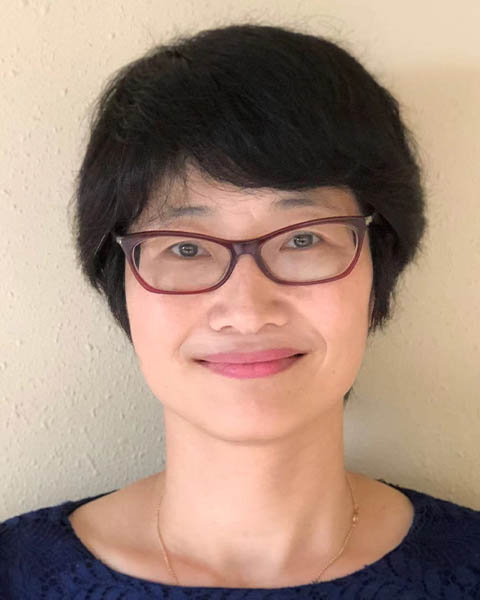}}]{Ying Tan}
is an Associate Professor and Reader in the Department of Electrical and Electronic Engineering (DEEE) at the University of Melbourne, Australia. She received her Bachelor’s degree from Tianjin University, China, in 1995, and her PhD from the Department of Electrical and Computer Engineering National University of Singapore in 2002. She joined McMaster University in 2002 as a postdoctoral fellow in the Department of Chemical Engineering. Since 2004, she has been with the University of Melbourne. She was awarded an Australian Postdoctoral Fellow (2006-2008) and a Future Fellow (2009-2013) by the Australian Research Council. Her research interests are in intelligent systems, nonlinear control systems, real time optimisation, sampled-data distributed parameter systems and formation control.
\end{IEEEbiography}

\vspace*{-2.5\baselineskip}

\begin{IEEEbiography}[{\includegraphics[width=1in,height=1.25in,clip,keepaspectratio]{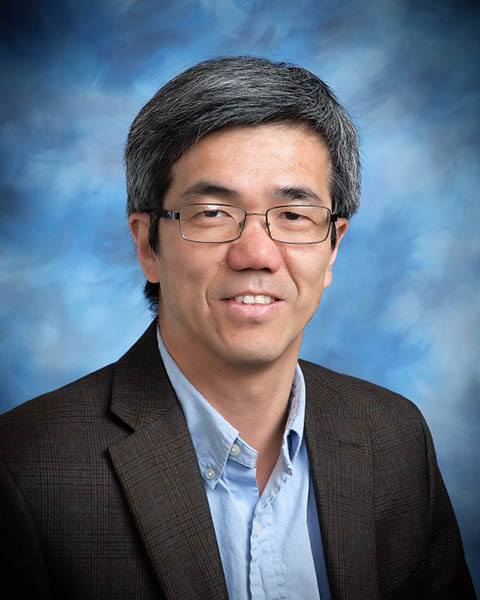}}]{Xiang Chen}
(M’98) received the M.Sc. and Ph.D. degrees from Louisiana State University, Baton Rouge, LA, USA, in 1996 and 1998, respectively, both in systems and control.

Since 2000, he has been with the Department of Electrical and Computer Engineering, University of Windsor, ON, Canada, where he is currently a Professor. His research interests include robust control, vision sensor networks, vision-based control systems, networked control systems, and industrial applications of control theory.
\end{IEEEbiography}

\vspace*{-2.5\baselineskip}

\begin{IEEEbiography}[{\includegraphics[width=1in,height=1.25in,clip,keepaspectratio]{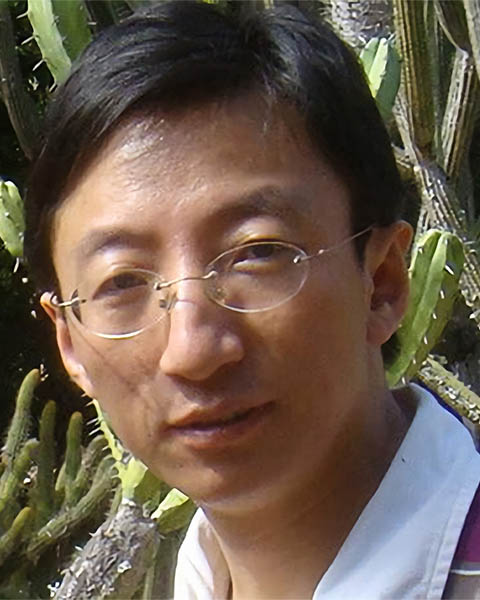}}]{Li Chai}
(S'00-M'03) received the B.S. degree in applied mathematics and the M. S. degree in control science and engineering, both from Zhejiang University, China, in 1994 and 1997 respectively, and the Ph.D. degree in electrical engineering from Hong Kong University of Science and Technology in 2002.

In September 2002, he joined Hangzhou Dianzi University, China. He worked as a postdoctoral research fellow at the Monash University, Australia, from May 2004 to June 2006. In 2008, he joined Wuhan University of Science and Technology. He is a professor at Zhejiang University since March 2022. He has been a visiting scholar at Newcastle University, Australia, and Harvard University. His research interests include distributed optimization,  filter bank frames, graph signal processing, and networked control systems.

Professor Chai is the recipient of the Distinguished Young Scholar of the National Science Foundation of China. He is currently an associate editor for the Decision and Control.
\end{IEEEbiography}







\end{document}